\documentclass[nonacm, acmlarge,screen]{acmart}

\AtBeginDocument{%
  \providecommand\BibTeX{{%
    \normalfont B\kern-0.5em{\scshape i\kern-0.25em b}\kern-0.8em\TeX}}}

\usepackage{graphicx}
\graphicspath{ {./Images/} }

\usepackage{cleveref}
\usepackage[disable]{todonotes}
\usepackage{caption}
\usepackage{subcaption}

\begin{document}

\title{Pragmatic auditing: a pilot-driven approach for auditing Machine Learning systems}  %

\author{Djalel Benbouzid}
\email{djalel@argmax.ai}
\affiliation{%
  \institution{Machine Learning Research Lab, Volkswagen Group}
  \country{Germany}
  \city{Munich}
}

\author{Christiane Plociennik}
\email{christiane.plociennik@dfki.de}
\affiliation{%
  \institution{German Research Center for Artificial Intelligence (DFKI)}
  \country{Germany}
}

\author{Laura Lucaj}
\email{laura.lucaj@tum.de}
\affiliation{%
  \institution{Technical University of Munich (TUM)}
  \country{Germany}
}

\author{Mihai Maftei}
\email{mihai.maftei@dfki.de}
\affiliation{%
  \institution{German Research Center for Artificial Intelligence (DFKI)}
  \country{Germany}
}

\author{Iris Merget}
\email{iris.merget@dfki.de}
\affiliation{%
  \institution{German Research Center for Artificial Intelligence (DFKI)}
  \country{Germany}
}

\author{Aljoscha Burchardt}
\email{aljoscha.burchardt@dfki.de}
\affiliation{%
  \institution{German Research Center for Artificial Intelligence (DFKI)}
  \country{Germany}
}

\author{Marc P. Hauer}
\email{hauer@cs.uni-kl.de}
\affiliation{%
  \institution{Algorithm Accountability Lab, TU Kaiserslautern}
  \country{Germany}
}

\author{Abdeldjallil Naceri}
\email{djallil.naceri@tum.de}
\affiliation{%
  \institution{Munich Institute of Robotics and Machine Intelligence (MIRMI), Technical University of Munich (TUM)}
  \country{Germany}
  \city{Munich}
}

\author{Patrick van der Smagt}
\email{}
\affiliation{%
  \institution{Machine Learning Research Lab, Volkswagen Group}
  \country{Germany}
  \city{Munich}
}

\begin{abstract}

The growing adoption and deployment of Machine Learning (ML) systems came with its share of ethical incidents and societal concerns.
It also unveiled the necessity to properly audit these systems in light of ethical principles. For such a novel type of algorithmic auditing to become standard practice, two main prerequisites need to be available: A lifecycle model that is tailored towards transparency and accountability, and a principled risk assessment procedure that allows the proper scoping of the audit.
Aiming to make a pragmatic step towards a wider adoption of ML auditing, we present a respective procedure that extends the AI-HLEG guidelines published by the European Commission. Our audit procedure is based on an ML lifecycle model that explicitly focuses on documentation, accountability, and quality assurance; and serves as a common ground for alignment between the auditors and the audited organisation.
We describe two pilots conducted on real-world use cases from two different organisations and discuss the shortcomings of ML algorithmic auditing as well as future directions thereof.

\end{abstract}


\maketitle

\section{Introduction}

Pushed by the increased adoption of Machine Learning (ML) methods, the topic of ethical and trustworthy artificial intelligence (AI) is getting more and more attention. The apparent ethical concern of deploying these methods in sensitive applications has led many organisations to publish their own guidelines on ethical AI, or similar codes of conduct. However, it has been shown that ethical guidelines have little influence on the actual decisions developers make \cite{McNamara2018Does, Whittlestone2019, Mittelstadt_2019, Hagendorff_2020}. This is caused by the fact that such guidelines are mostly too abstract and technical solutions to support them are  often missing, encouraging developers to a cession of responsibility.

A recently discussed solution to the problem lies in adopting the process of auditing an organisation or its processes when designing, developing, and operating systems with ML components -- we call them ML systems henceforth. Auditing these systems can increase accountability and help operationalize ethics guidelines \citep{brundage2020toward}. Audits can be performed internally, i.e., by members of the same organisation, or by a third-party auditor. The latter is considered especially useful for higher-risk applications.

Albeit costly, third-party auditing leads to greater apparent trust by the different system stakeholders, especially if the audit leads to an acknowledged certification \citep{knowles2021sanction}. However, internal audits also have their merits as they incentivise the production of data and model documentation, risk analyses, and all sort of artefacts that increase the traceability of the audited system.

Auditing, as an institutionalised practice, however comes with certain conditions of feasibility, without which the auditor and the audited organisation cannot align on the expectations nor on the practices to be implemented. In particular, audit specifications have to be defined and known by all the parties.
These specifications include the standards to be implemented, the artefacts and evidence to be collected, the tests to be conducted, as well as the roles that ought to be defined and interviewed within the audited organisation.
Furthermore, auditing requires the ability to assess risks and further implement the  corresponding control measures.
As standard specifications in the challenging domain of assessing risks of ML systems are lacking, a different approach is needed.
We propose in this paper a procedure inspired by Information Systems auditing, which we alter to account for the aforementioned shortcomings. We introduce a high-level lifecycle model that encodes ethical requirements such as transparency and accountability and translates these into systemic practices. The lifecycle model serves as a common map between the auditor and the auditee and sets the expectations as for what precisely should be audited and how.
The lifecycle model further allows us to define a risk assessment scheme based on the EC ALTAI \cite{hleg2020assessment}, which allows to narrow the
scope of auditing down to major ethical concerns.
Finally, we illustrate the proposed procedure with two pilots conducted within the members of the etami (Ethical and Trustworthy Artificial and Machine Intelligence) consortium \cite{etamieu} , unveiling possible improvements and future directions.

\section{Related Work}

As ML systems enter many areas of everyday life, questions about the impact of their performance are becoming highly relevant~\cite{whittlestone2019ethical}.
Ethical challenges with the deployment of ML systems are widely recognised~\cite{char2018implementing, milano2020recommender, lange2021combating}.
For instance, it has been shown that ML systems can produce predictions that disproportionately affect vulnerable minorities and contribute to the re-enforcement of the structural bias already present in society~\cite{o2016weapons, buolamwini2018gender, raji2020closing, kirkpatrick2017s, amini2019uncovering}.
Nonetheless, few have provided guidance on how to address such issues in practice. Effective, actionable and comprehensive methodologies to understand, address and mitigate such impact remain under-investigated.
Auditing is considered one of the major mechanisms for supporting verifiable claims and converging towards trustworthy ML systems~\cite{brundage2020toward}.
Algorithmic auditing of ML systems in particular has gained a considerable recognition as an opportunity to harness the potential of ML models as well as detecting and mitigating the problematic patterns and consequences of their deployment in sensitive decision-making contexts~\cite{raji2019actionable, koshiyama2021towards, wiens2020diagnosing}.

\subsection{Literature on algorithmic auditing}
Several practices that need to be embedded at specific phases of the development of an ML model have been explored. For instance, audit trails, verification and bias testing as well as explainable user interfaces have been proposed to be applied and adapted to auditing procedures~\cite{shneiderman2020bridging}.
\cite{shneiderman2020bridging} presents a large number of recommendations related to different levels of governance, namely teams, organisations, and industry. However, it provides little guidance on how to implement such measures in practice.
Others have examined templates for fundamental documentation practices. Similar to the supplier’s declaration of conformity (SDoC), which provides information on how a product conforms to the technical standards or regulation enforced in the country it is deployed, FactSheets have been proposed. They constitute a comprehensive documentation framework~\cite{arnold2019factsheets,richards2020methodology}.
Such documentation aims to record the practices employed in the development of an AI model and discloses its exact intended purpose. The aim is to increase the consumer's trust and to address the potential ethical concerns emerging in this phase~\cite{arnold2019factsheets,richards2020methodology}.

Such practices constitute essential steps that have to be embedded into auditing procedures. However, on their own, they are not sufficient to address all the challenges of the impact assessment of ML systems. \cite{koshiyama2021towards} provides a detailed overview of auditing tools for AI systems.
The authors tackle different ethical requirements such as fairness and explainability and suggest mitigation measures to adhere to ethical standards. The paper is rather comprehensive in its coverage. However, it does not provide a general procedure that brings consistency to all components it discusses.
A major contribution towards defining a comprehensive framework for auditing ML systems was proposed by~\cite{raji2020closing}, wherein a framework for internal auditing is introduced to enable proactive interventions. The framework tackles the practices necessary along the pipeline of an ML model to record important design decisions and to identify the causal relation between such decisions and the risks that might emerge and relate to ethical failures.
This process allows stakeholders of the audit to define their own ethical principles to audit against before the deployment. It is debatable whether such a procedure can discover whether a system being audited is really ethical. \cite{raji2020closing} target their approach explicitly towards internal auditing and towards a certain lifecycle phase. We believe an actionable audit procedure should be suited for both internal and external audits -- the difference is explained in~\cref{appendix:int_ext_audit} -- and for any lifecycle phase.

\subsection{Toolkits}

Auditing ML systems is often seen through the lens of a single ethical concern, such as discrimination and bias. To enable translating ethical principles into practice, several toolkits have been designed. They aim to detect and understand the potential sources of bias and unwanted consequences caused by the deployment of ML algorithms. Hence, a number of auditing tools and frameworks tackle specific areas, e.g.\  diversity \cite{diversitywork}, bias in datasets~\cite{saleiro2018aequitas,bellamy2018ai}, or explainability~\cite{nori2019interpretml,arya2020ai}. These tools and frameworks complement the overall procedure we describe in this paper.
Certain toolkits focus on risk assessment \cite{ethicstool}. Others provide guiding measures to proactively understand, identify and address the sources of bias and discrimination that disproportionately affect vulnerable minorities in society \cite{diversitywork}. Other toolkits aim to guide the developers to integrate ethical procedures into AI systems during the development and operationalisation phase, e.g., via simple methods such as questionnaires \cite{aiblindspot, dataethics}. In general, such toolkits can play a significant role in simplifying the procedure of translating ethical principles into practice. They help developers to foresee potential sources of bias as well as the impact their system might have on the well-being of individuals. However, they provide guidance on a task-specific basis. They are therefore insufficient for addressing all the issues that can potentially emerge through the entire pipeline of the design, development and deployment of an AI system.

\subsection{Guidelines and regulation in the EU}

In the last few years the European Commission (EC) has proposed guidelines on AI such as the Ethics Guidelines for Trustworthy AI in 2019~\citep{ai2019high} and the White Paper on AI in 2020~\citep{european2020white}.
Recently, the EC has worked on translating the principles proposed by such guidelines into actionable practices through tools, such as the Assessment List for Trustworthy Artificial Intelligence (ALTAI). Its aim is to help developers to assess the impact of their AI systems~\citep{hleg2020assessment}.

In 2021, the EU Artificial Intelligence Act has been issued, a proposal for legislating AI systems~\citep{european2021proposal}. Such proposal is the result of the significant work, conducted by European legislators to guarantee the development of safe AI systems conforming with the human-centric values of the European Union.
The proposal characterises some AI applications as high-risk and enforces an \emph{ex ante} validation process prior to their marketing.

\subsection{Discussion}
In spite of this significant work, the current practices proposed in the field of algorithmic auditing present several gaps and are generally hard to apply in practice, especially at an organisational scale.
Unlike traditional software systems, ML algorithms require new methods for system testing \citep{jockel2021towards}, novel ways and processes for assessing their quality and conducting their audits.
Yet the current literature provides little guidance on the organisational practices that could provide a pragmatic answer to these challenges.

\section{The audit procedure}
\label{sect_contributions}

\begin{figure}
\centering
  \centering
  \includegraphics[width=.7\linewidth]{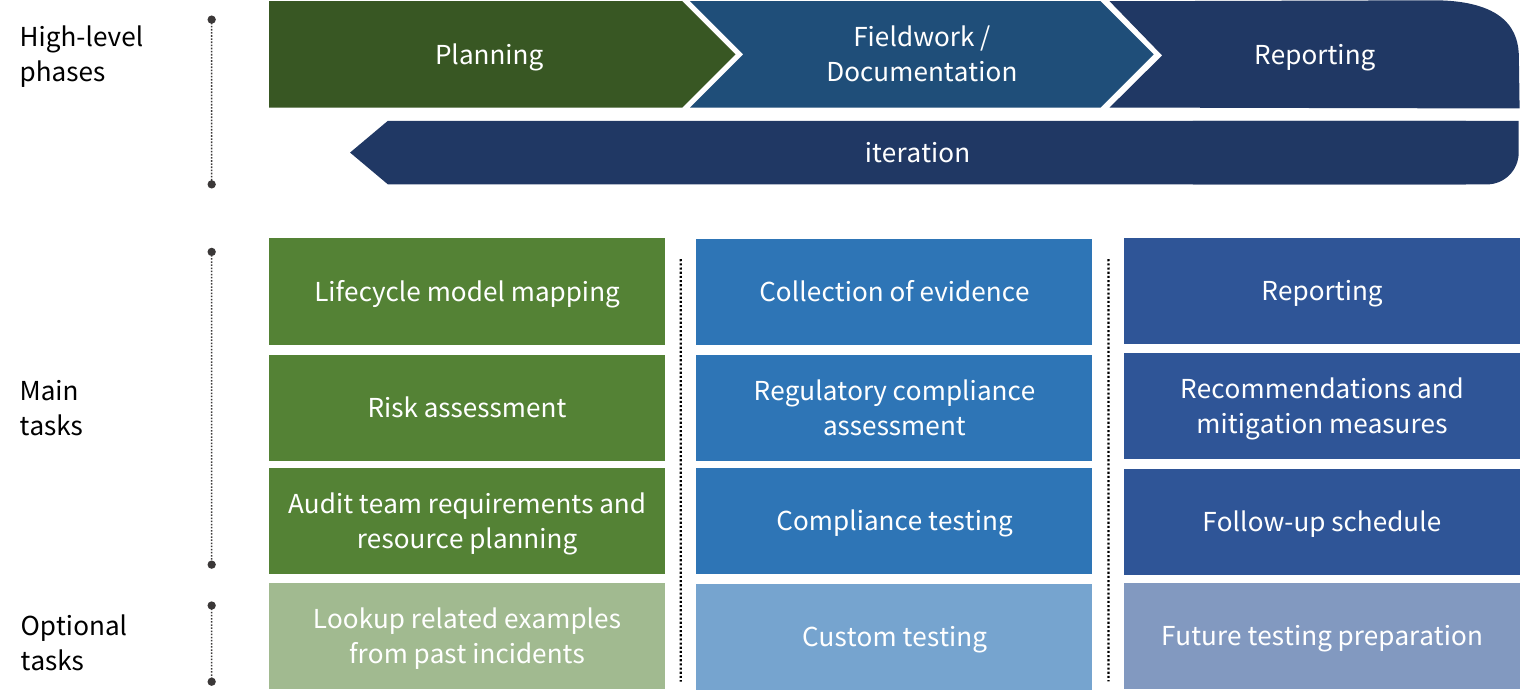}
  \captionof{figure}{A high-level view of the audit process.}
  \label{fig:audit_process_short}
  \Description{The image describes a process in three consecutive steps: first, planning, then fieldwork and documenting, and finally reporting.}
\end{figure}

As machine learning systems often coexist with larger information systems that are likely already subject to some form of auditing, we propose a procedure inspired by the ISACA framework \cite{isacaorg}, a widely adopted standard in Information Systems auditing.
Our aim is to allow machine learning auditing to be easily integrated into existing organisational practices, whether they are based on internal or external assessments\footnote{We discuss the the differences between internal and external audits in \cref{appendix:int_ext_audit}.}.
Our framework, summarised in \cref{fig:audit_process_short}, consists of three phases, namely the planning, the fieldwork/documentation, and the reporting.
Although these three phases are usually depicted as sequential, we emphasise here that for ML systems the audit should be a continuous process, reflecting their fast and repeated development iterations.
In the following sections, we adapt each of these phases to the specifics of ML systems, describe the challenges thereof, and propose solutions. Note that we will only describe those tasks that are not self-explanatory.

\subsection{The planning phase} %
\label{sub:the_planning_phase}
This phase aims at setting the precise scope of the audit and a roadmap for the subsequent fieldwork and documentation, in the following called \emph{audit program}. Therein, the auditor looks up previous audit reports if available and conducts a risk assessment.
The audit planning also allows to specify the resources and skills needed for the audit; in particular, a multi-disciplinary background is often required for the auditing team, in order to properly cover the complexity of deployed ML systems.

A key element of this phase is a good understanding of the overall architecture and the processes underlying the design, development, and deployment of the ML models.
Following standard practice, we enable this by following a lifecycle model. The lifecycle model should reflect ethical principles, highlight the major corresponding risks, and allow a straightforward assessment of the risk mitigation measures.
Below, we propose a lifecycle model that satisfies these requirements.

\subsubsection{Lifecycle model mapping} %
\begin{figure}
  \centering
  \includegraphics[width=.8\linewidth]{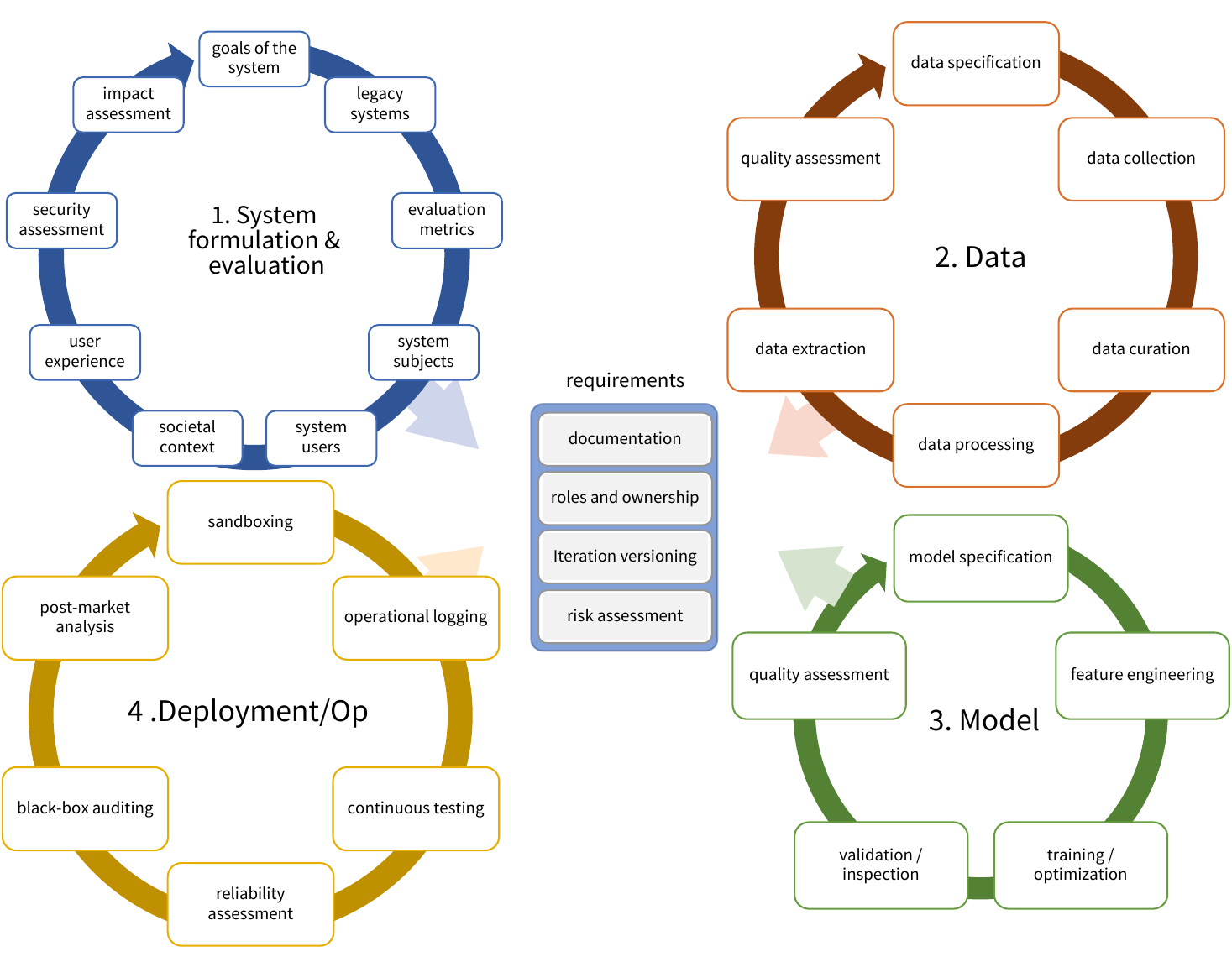}
  \captionof{figure}{A lifecycle model for ML systems design to align the auditor with the auditing organisation. Each of the steps can be subject of a required documentation, a proper versioning, and a clear definition of ownership and roles. Furthermore, the lifecycle model is accompanied by a risk-assessment procedure that relates to each of its steps.}
  \label{fig:lcm}
  \Description{The image describes a lifecycle cycle model for Machine Learning systems. It consists in a circular process with four steps, each of which contains a circular subprocess.}
\end{figure}

\label{ssub:lifecycle_model_mapping}
Serving as a common map between the different parties throughout the audit, a lifecycle model is key for scoping and understanding the overall architecture of the system.
Most of the lifecycle models in the literature, however, put a heavy focus on the technical aspects of the system~\citep{ashmore2021assuring} and fail to translate the ethical principles such as accountability and transparency into systemic practices.
We argue that, for an easy and transparent implementation of these principles, the lifecycle model of the ML system itself has to be interrogated. We therefore propose an improved lifecycle model, shown in \cref{fig:lcm}.
Our lifecycle model lies on a high enough abstract level to interface with existing architectures and covers the four major steps of ML systems: the formalisation, the data management, the model management, and the deployment; however, it differs in that it highlights three fundamental aspects that foster accountability and transparency:
\begin{description}
\item{\sl The agility of each phase.} Often, ML lifecycle models are depicted as a linear process spanning from the business requirements and data management, all the way to the model training and deployment. Despite the common emphasis on the possibility to backtrack into a previous phase at any moment, these models fail to depict a process centred around quality.
Instead, we highlight the agile nature of each of the four phases and stress the necessity to
specify before implementing, document while making, and the constant assessment for quality.

\item{\sl Designing around transparency and accountability.} While common lifecycle models are constructed around technological and functional boundaries, we further align the different phases and steps of our lifecycle model based on the roles and ownership that are required for designing ML systems. These roles are associated with the relevant documentation that is expected to be produced as well as a clear and justified go/no-go scheme for each of the critical phases.

\item{\sl The continuous impact assessment.} It is commonly accepted that during the design of a high-risk application, a multitude of variables are to be taken into account, such as the specificity of the  targeted sector, a thorough understanding of the end users and the subjects of the system, the risks induced by the ML model itself, etc. Our lifecycle model makes that explicit and further emphasises a continuous impact assessment using the data collected after deploying the model.
\end{description}

\noindent
Mapping the audited system onto the lifecycle then simply translates into selecting the relevant phases and steps and defining the corresponding artefacts to be collected. As we describe below, this short-listed set of steps also allows for risk assessment to be tailored and conducted.

\subsubsection{Risk assessment} %
\label{ssub:risk_assessment}

Auditing generally assumes having access to a knowledge base of documented risks, either gathered from prior incidents or built in a more proactive and preventive manner.
Such a tool, however, is difficult to obtain for ML systems~\citep{boyarskaya2020overcoming}. Instead, we introduce a risk-assessment method that relies on the previously introduced lifecycle model. We break the risk analysis down according to the steps of the lifecycle, hence reducing the complexity of overall endeavour down to specific components of the system.
Each component can then be further documented with its associated risk.
This approach has the additional merit of making knowledge of risks transferable across projects or even organisations, as long as they share similar steps in their lifecycle.
To kickstart this risk-assessment methodology, we resort to the work of the European Commission, namely the Assessment List for Trustworthy Artificial Intelligence (ALTAI). This assessment tool allows us to define the relevant questions to raise during each of the phases of the lifecycle. It not only enables fieldwork and documentation later on but also helps to define the appropriate mitigation measures.
We provide a comprehensive lifecycle-based assessment in~\cref{appendix:risk_assess}.

\subsection{The fieldwork and documentation phase} %
\label{sub:the_fieldwork_and_documentation}
Having produced the audit program during the planning phase,
the auditor can start to acquire the actual data for the audit,
the \emph{evidence}, in order to verify the compliance with the relevant regulation, the potential standards, and most importantly, to evaluate the control measures put in place through a series of tests.

\subsubsection{Collection of evidence} %
\label{ssub:collection_of_evidence}
Evidence can be based on two kinds of mechanism: transparency mechanisms (disclosed information) and examinability mechanisms (experiments based on granted access)~\cite{hauer2022overview}. Transparency mechanisms allow an external auditor to assess information made transparent by the developers (e.g., with datasheets~\citep{gebru2021datasheets}, factsheets~\citep{arnold2019factsheets,richards2020methodology}, model cards~\citep{mitchell2019model}, etc.).  The auditors need to rely on the correctness of the information collected in the documentation phases in the lifecycle model. To this day, no standardised documentation process has been established.
With examinability mechanisms (e.g., access to the database, ability to issue queries to an ML component and directly inspect the outcome, etc.)\ though, the auditor can validate information previously made transparent and also directly conduct experiments on the system to be audited. As examinability mechanisms are based on access to the system granted to an external entity, this is not made explicit in the lifecycle model, as the accesses can be created post-hoc, on a per-need basis.
Examinability mechanisms can be used to investigate the data management, model management, and deployment/operationalisation phases.

\subsubsection{Compliance testing} %
\label{ssub:compliance_and_substantive_testing}
Collecting evidence allows the auditor to conduct so-called \emph{compliance} testing. These are all the tests that assess the potential discrepancies between the specifications collected during the formalisation of the ML system and the their actual implementation~\cite{guerin2003verification}. This includes, for instance, insufficient documentation of the data collection process, insufficient explanations to the final users, or a quality assessment based on unfitting metrics, as there are more than 20 different operationalisations of ML quality (frequently called \emph{performance measures}) and countless variations and combinations, contradicting each other to a great extent~\cite{sokolova2009systematic} (the same holds for more specific properties like fairness~\cite{verma2018fairness} or diversity~\cite{tang2006analysis}).
Compliance testing plays an important role in ensuring the quality of ML systems, thus fulfilling one of the major objectives of auditing.

As the developing team of a system most likely have their own definition of quality, an auditing procedure could focus on validating whether the system complies to this definition of quality and whether this definition is reasonable. Not only this remains challenging due to the aforementioned reasons but more crucially, such an approach cannot unveil blind spots in the design of these systems, which is another major objective of auditing. One approach to solve this problem, as suggested by~\cite{hauer2021assuring}, is to let the developers build a structured argumentation about why they consider their system having a sufficient quality to be used~\cite{hauer2021assuring}. Then, the auditing procedure can more easily challenge that argumentation, the explicit assumptions it is based upon and the provided evidence.

\subsubsection{Custom testing} %
Especially with emerging technologies it is inevitable that testing procedures lack thought due to inexperience (for instance, COMPAS has shown that fulfilling given specification may not be sufficient, e.g., due to a bad selection of performance measures~\cite{Angwin2016a}). If the auditing team recognises such a situation, they can define their own tests. This way they can test for aspects and in ways the developers didn't. It should be noted that this can lead to audits becoming incomparable to a certain degree.

\subsection{The reporting phase} %
\label{sub:the_reporting_phase}
As the auditor compiles the results of the different tests as well as the assessment of the implemented controls, they also define the criteria for a new round of auditing. This can correspond to a periodic schedule or to an event-based procedure, such as the deployment of the system in a new social context, or the accumulation of negative feedback by the users of the system (or its subjects, as the latter can be part of a system upon which they have no agency). Importantly, the auditor must ensure that the mitigation measures recommended in consequence of the audit should be followed and implemented before the subsequent iterations.

\paragraph{Preparing for future testing.}
As auditing under test conditions may not fully reflect the system behaviour in application (e.g., due to artificial test data or a system that is blocked for inputs from outside the auditing procedure, etc.), it makes sense to implement test procedures to be executed on a regular basis during application and inspect the collected results in the next auditing iteration. To further improve the process, the documentation can be limited to failing tests, error aggregations (like an error rate) or the the computation of carefully selected performance measures, based on batches of inputs (e.g., after $n$ inputs, compute for the last $n$ outputs the fairness measure of conditional independence~\cite{barocas2018fairness} with regard to gender fairness and document if it is below $95\%$).

\section{Conducting the Pilots}

The audit procedure proposed in this paper is the result of an ongoing process of conducting pilots within etami \cite{etamieu}.
Hence, the aim is not to provide a ``fit-all'' solution for auditing, but rather to advance the conversation around the best practices in the field.
Below we illustrate the audit procedure on two use cases.

When conducting the pilots, we did not audit all the stages in the lifecycle model, but only those that made sense or were auditable. For example, the system described in \cref{sect_motor} is almost fully developed and close to market entrance. When auditing it, we found it was not possible to get access to the actual data and all the models employed during development because some of these artefacts were \emph{not accessible anymore}. On the contrary, GARMI (described in \cref{sect_garmi}) is an early-stage research project. The relevant data and models are \emph{not yet accessible}. Hence, we colour-coded the stages in the lifecycle model that we audited in blue, the stages we did not audit because we found them not relevant in yellow and the stages we could not audit due to missing data/artefacts in white (see \cref{fig:lcm_motor}).

This indicates two things: First, it is possible to take a pragmatic approach towards auditing. It is not necessary to audit each and every phase of the lifecycle model. Second, the proposed colour-coding can serve as an indicator of ``audit coverage'': The more phases are blue, the more comprehensive is the audit. This can be viewed as an analogy to test coverage in unit testing.

\subsection{Auditing an AI-assisted calibration system}\label{sect_motor}

This use case, provided by a first tier supplier, aims at automatising the calibration of a complex safety component. Traditionally, this calibration is done manually by trained engineers. The aim of the project is to keep the engineers in the loop, and to merely provide them with a recommended calibration.
This audit pilot was conducted jointly with the product owner, the lead ML developer, and the lead UX designer.

\paragraph{System description:}
The ML system outputs the calibration of a complex safety-relevant physical system, which must be fully operational under a variety of environmental conditions. This parametrisation and calibration, traditionally done by domain experts, is a highly complex procedure that should guarantee the successful and safe usage of the actuated object in any circumstance, i.e., depending on the ambient temperature, whether the object (usually an electrical actuator acting on a heavy object) is on a slope, etc.
The ML model thus captures the historical knowledge accumulated by the domain experts and serves as a recommendation engine:
the users of the system -- the engineers -- are presented with a list of recommended parametrisations ranked by performance. They still have to choose one of the recommendations and manually apply it to the motor.

\paragraph{Audit planning and lifecycle model mapping:}

\begin{figure}
  \centering
  \includegraphics[width=.8\linewidth]{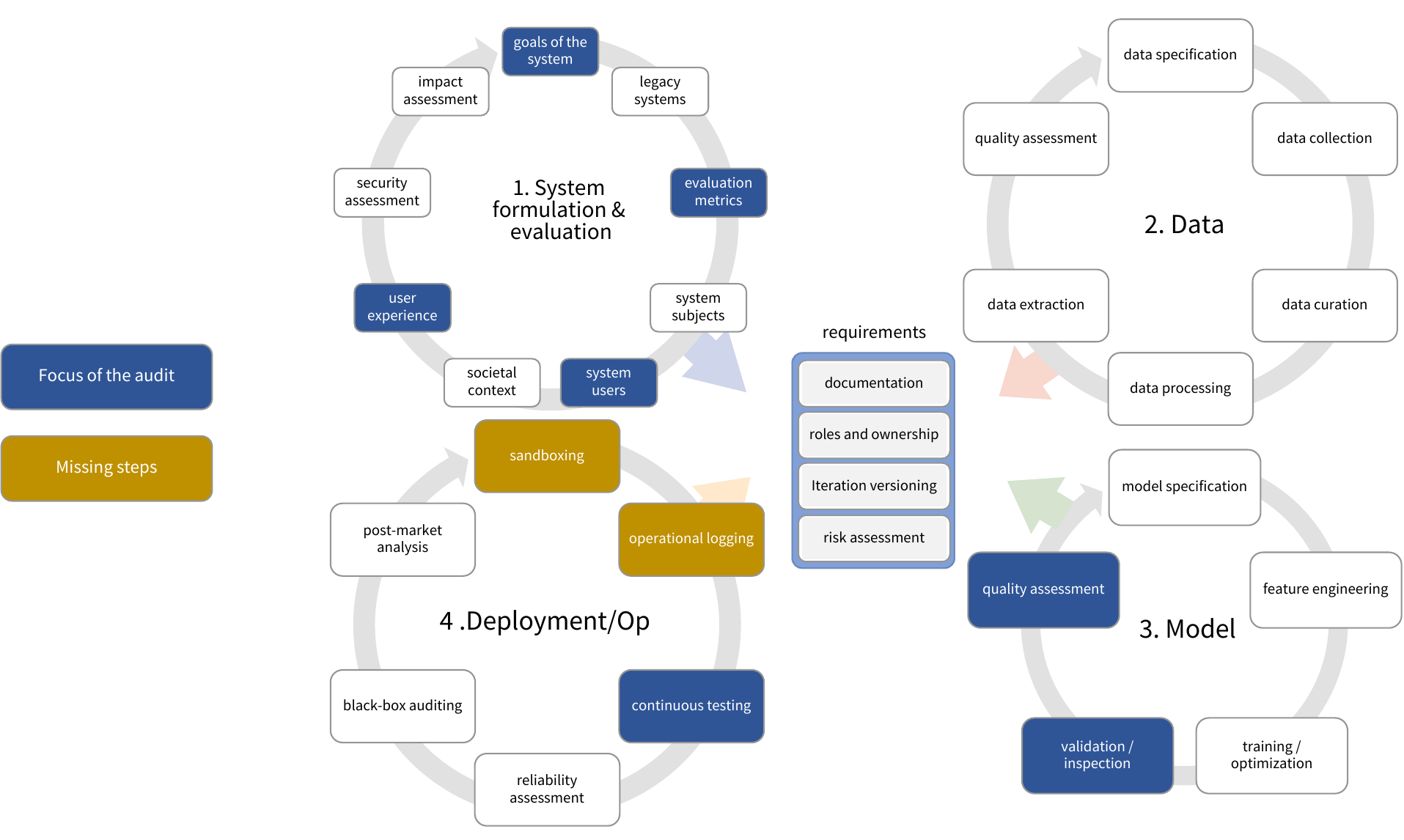}
  \captionof{figure}{The focus of the first pilot. For each of the blue steps, a thorough inspection of the documentation, roles, and ownership was conducted. These steps also served narrow the scope of the risk-assessment. The yellow steps were highlighted as expected but missing.}
  \label{fig:lcm_motor}
  \Description{}
\end{figure}

For this particular pilot, we did not cover the whole lifecycle model described in \cref{ssub:lifecycle_model_mapping}. We rather put the emphasis on the formalisation, the model management, and the operationalisation phases. The rationale is that the data management pipeline was part of the legacy infrastructure and thus already subject to substantial scrutiny. Furthermore, the system did not collect new data and was already compliant with the General Data Protection Regulation.
Nonetheless, as expected, the lifecycle model in its entirety helped drive the initial discussions with the auditee, serving as a map to understand the different system components and overall architecture.
Mapping the scope of the audit to specific steps of the lifecycle allows to define the expected documentation and specifications to collect and the clear attribution of roles and ownership, as depicted in \cref{fig:lcm_motor}. It also narrows down the risk-assessment procedure to these specific steps and helps highlight the related ethical concerns.

\paragraph{Risk assessment:}

Having defined the scope of the audit based on a subset of the lifecycle model, the risk assessment can then be conducted. The procedure consists in using the  database of questions defined in \cref{appendix:risk_assess} and only retaining those relevant to the selected steps of the lifecycle model. This can be easily implemented in a simple tool. For the sake of this illustration, we resort to a spreadsheet and use its filtering capabilities.
For instance, given that the UX is part of the system formulation, filtering out the database described in \cref{appendix:risk_assess} allows to  highlight important questions such as,

\begin{itemize}
  \item Did you explain the decision(s) of the AI system to the users?
  \item Do you continuously survey the users to assess whether they understand the decision(s) of the AI system?
  \item Did you provide appropriate training material and disclaimers to users on how to adequately use the AI system?
\end{itemize}

The answers to these questions, whether provided by the auditee or directly collected by the auditor in some cases, allow to define the potential ethical concerns and consequently the mitigation measures to address them.

In particular, after applying the aforementioned procedure to the short-listed lifecycle, the following ethical concerns were highlighted,

\begin{itemize}
  \item privacy,
  \item transparency and explainability, and
  \item robustness and safety.
\end{itemize}

Below we detail these concerns and discuss the mitigations that an auditor could provide.

\paragraph{Privacy}

The system is designed to be used by engineers locally and its usage is not intended to be scaled up. Furthermore, the system does not collect further data about its users and only relies on a legacy dataset consisting in expert rules. Despite containing identifying records, data privacy is not an ethical concern in the present case as the system does not collect further data from its users and already complies with the regulation.

\noindent\textbf{Recommendation:} none

\paragraph{Transparency and explainability}

The system is transparent in that it provides a detailed trace of the overall procedure and the output is clearly displayed to the user. Furthermore, the outputs -- in this case the list of parametrisations -- are ranked by performance showing the best recommendations at the top.
As the system is aimed at being mainly a support for the engineering and not as a replacement, however, the interface does not clearly convey the possible uncertainty about the outputs, nor that their usage should remain subject to the expert’s supervision.

\noindent\textbf{Recommendation}:
To be a full “human-in-the-loop” system, it would be informative to assess the over-reliance of the users on the system’s output. Two ways to achieve this are the following.
\begin{itemize}
  \item Logging the chosen parameters and offline analysis: By logging what parameters the operator selects (which was not done at the time of the audit), one can validate this choice a posteriori, either by running the optimisation procedure anew or with a manual check by other operators.
  \item Randomising the displayed parameters: “Dry-run” experiments can be set up wherein the operator is presented with a sequence of parametrisations that are either randomly permuted or slightly re-ordered. Such an experiment would potentially unveil whether the operators over-rely on the system’s output even when better parametrisations exist.
\end{itemize}

\paragraph{Robustness and safety}

The design choices, the near-misses, as well as the investigation conducted on the failures, have in general to be documented. The developers of the ML system can document the design input, output, review, verification and validation, as well as the transfer and changes in order to attempt tracing back what might have caused the adverse effect of the system in a given circumstance (record design history). Post-market surveillance is also essential for safety, as the near-misses have to be constantly monitored.

\medskip\noindent\textbf{Recommendation}:
For the present use case, an additional documentation should include the thresholds for the acceptable performance of the system; for instance, how the system is supposed to perform given the different circumstances and which risks can be considered acceptable.
Once the system is in beta version, conducting adversarial testing focusing on the edge cases and seeing the performance and failure possibilities could be fundamental in assuring safety.

\subsection{Auditing the vision system in a Geriatronics project }\label{sect_garmi}

\begin{figure}[!t]
\centering
    \includegraphics[width=0.5\columnwidth]{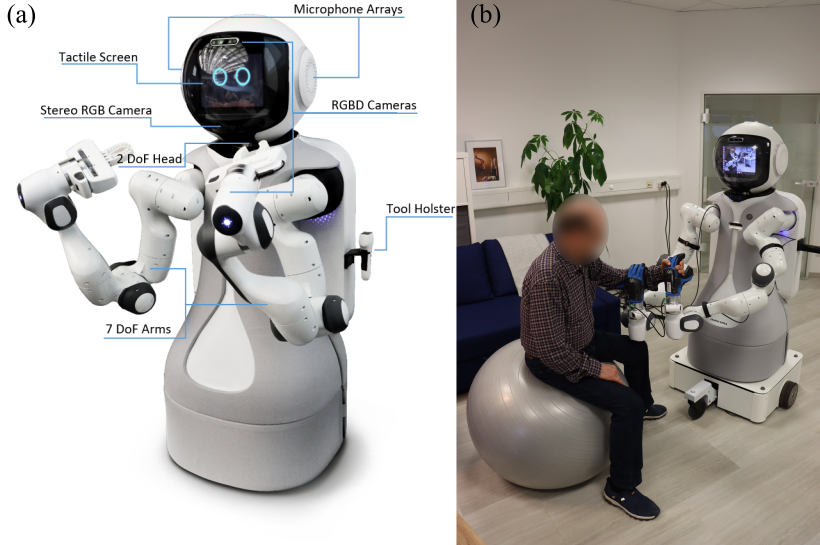}
    \caption{GARMI: service humanoid robot \cite{Garmi_2021}. (a) Description of GARMI's different modules. (b) GARMI helping an elderly person in rehabilitation scenario.}
  \label{fig:garmi}
\end{figure}

This use case is provided by GARMI, a service robotics platform dedicated to assisting elderly people in their daily life activities \cite{Garmi_2021} (\cref{fig:garmi}).
GARMI provides manual and semi-autonomous assistance and multimodal interaction on the one side and at the same time an interface for telemedicine and telerehabilitation on the other side. Design and capabilities have been developed based on a participatory design process with elderly users and medical practitioners.
It includes the mechatronic platform and modules for human--robot interaction, mobile manipulation, perception modules (auditory, computer vision and haptics). This modular design enables rapid prototyping and field-testing of new behaviours.

\paragraph{System description:}

In this auditing pilot, we focused on GARMI's perception vision module component. It is used to perceive the surrounding environment (e.g., wall, table, etc.) as well as the elderly person during human--robot interaction scenarios. The computer vision module is based on RGB-D (colour and depth) cameras (\cref{fig:garmi}a). The latter allows to capture the patient's 3D human skeleton used for motion analysis by identifying the patient's joint link lengths %
and computing the relative pose of the patient, with whom GARMI will later physically interact.
Concurrently, the patient's facial expression is analysed and used as input for motor actions and safety features. One use case for GARMI is to help an elderly person with some specific rehabilitation scenario (\cref{fig:garmi}b). During this process, the patient's facial expression is captured via a video feed from GARMI's on-board cameras and analysed online with an ML algorithm. If this analysis classifies a facial expression that indicates pain, the rehab procedure is aborted and GARMI switches into standby state.

This pilot involved the participation of the project Y lead, a robotics researcher involved in the project, and the team of ethicists who participated in the original ethical assessment of the project.

\paragraph{Audit planning and lifecycle model mapping:}

GARMI is still a research product under development. Our initial audit iteration indicated that while the design of the system underwent a substantial ethical analysis, the data and model management still did not produce enough artefacts to make the whole system auditable. Hence, this pilot illustrates how early auditing can help increase the overall quality of the system while reducing potential compliance costs.

\begin{figure}
  \centering
  \includegraphics[width=.8\linewidth]{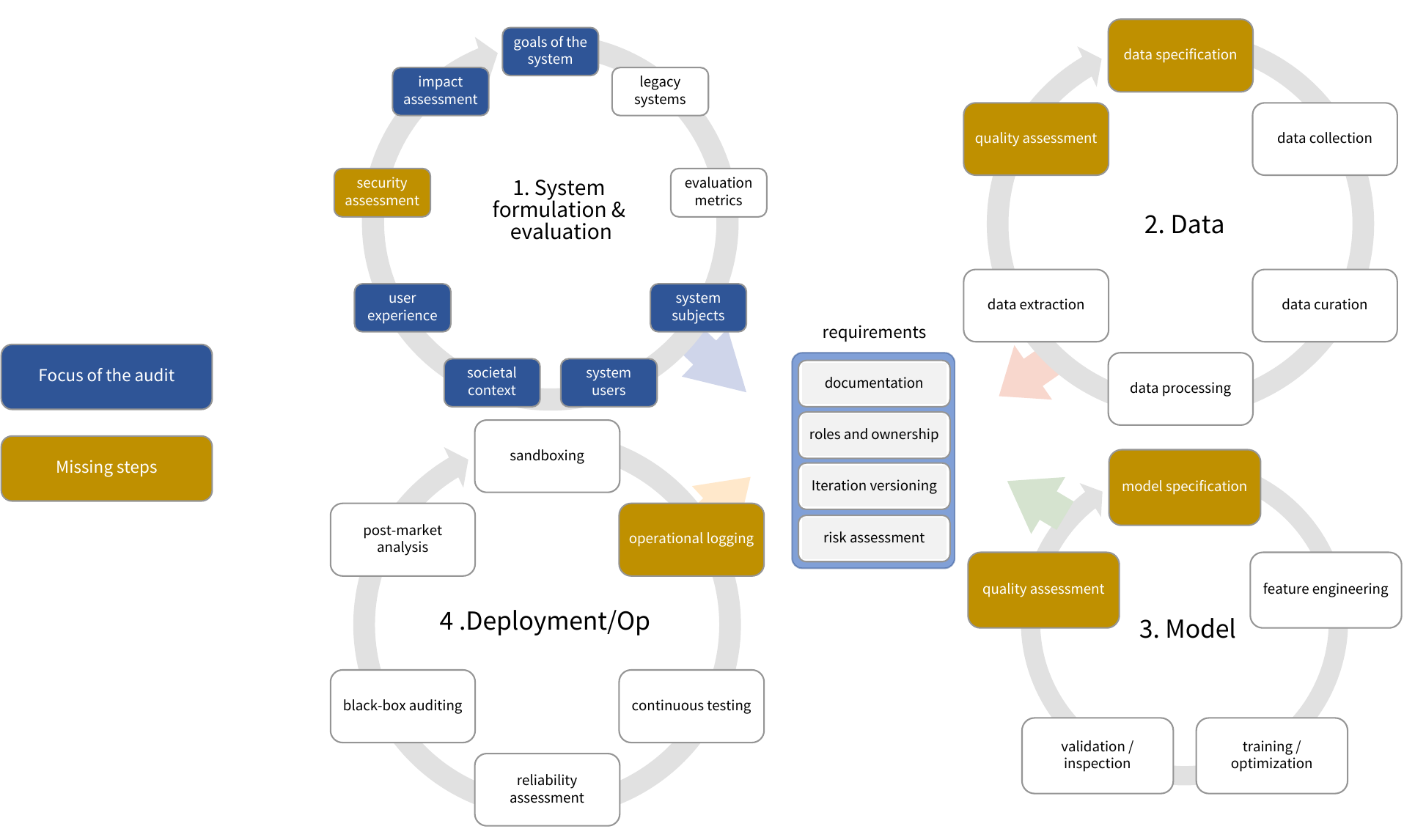}
  \captionof{figure}{The focus of the second pilot. For each of the blue steps, a thorough inspection of the documentation, roles, and ownership was conducted. These steps also served narrow the scope of the risk assessment. The yellow steps were highlighted as expected but missing.}
  \label{fig:lifecycle_garmi}
  \Description{}
\end{figure}

\paragraph{Risk assessment:}
Given the early stage of the project, several ethical requirements of the ALTAI were considered and assessed (cf. \cref{appendix:risk_assess} for the list of requirements and their corresponding questions).
This resulted in a comprehensive questionnaire that was exchanged with the audited organisation. Importantly, many of these requirements will be subject to further audit iterations.

\paragraph{Transparency and accountability:}
The audit helped identify the key elements to be documented and made traceable for a further audit. In particular, the data and model management should be precisely specified and quality-assessed during each iteration, as depicted in \cref{fig:lifecycle_garmi}. In this case, templates for documentation could be provided by the auditor, by tailoring datasheets \citep{gebru2021datasheets} and modelcards \citep{mitchell2019model} to this particular use case, along with a proper versioning for greater traceability.

\paragraph{Societal and environmental well-being:}

Within the Geriatronic project \cite{geria} there is already a focus on the planning and implementation of studies on ethical, social and legal issues on the topics and research areas focusing on service humanoids or telemedicine. Specifically, in this work, ethical, social and regulatory issues are taken into account in order to integrate them as part of a system intelligence \cite{McLennan2020, fiske_2020}.

\paragraph{Privacy and data governance:}

Several issues were raised during this early stage auditing process that the project lead has to address during next development round. For instance, the vision system module should explicitly demonstrate its compliance with General Data Protection Regulation, as it collects facial expression, gestures, and body skeleton data of the patient.

\paragraph{Prior ethical analysis:}

Overall, this system, while still in research phase, has already been the subject of a substantial body of work on the ethical and societal implications of Geriatronics \citep{McLennan2020}.
In particular, the project has been following the ``embedded ethics and social sciences'' approach
, which adopts a qualitative research framework and explores the research practices, activities and environments contributing to the development in a ``bottom-up'' manner (by using interviews, ethnography and focus groups) to come up with interventions later in the course of the development that are based on these insights. The aim is not to certify or audit technological systems, but to provide a more holistic account for how the research relates to wider society.
This work complements the more top-down approach of algorithmic auditing and provides further material for understanding the system and assess it.

During the subsequent audit iterations, it is planned to review the aforementioned ethical requirements anew and further put an emphasis on diversity, non-discrimination, and fairness, once enough evidence is produce thereof.

\section{Lessons Learned from the Pilots}

Piloting the audit procedure along with the lifecycle model provided us with substantial feedback that will serve further developments. Below we describe the lessons learnt along the way.

\paragraph{Auditability criteria:}
Not all systems can be audited, and so for various reasons.
For some systems, strong restrictions on data and code access are implemented, for others, the necessary artefacts, documentation, and logs are simply nonexistent.
Hence, there is a  need for a set of auditability criteria to define and check against prior to the audit planning.

\paragraph{No one-size-fits-all:}
Depending on the risk category of the ML systems, which in turn depends on variables such as the sector within which it is deployed, the level of automation, the potential risks for privacy etc., different kinds of audit procedures could be relevant.
The palette of procedures can for instance vary depending on the cost incurred by the audit itself and the compliance cost it yields. Some audits require a relatively shallow fieldwork while others involve a deeper look into the system. Some auditors require basic knowledge of machine learning systems while others need to define their own tailored battery of tests. It therefore appears relevant to distinguish the different types of audit along with their conditions.

\paragraph{Continuous auditing:}
Auditing is often conducted before market deployment or even after the development of an ML system. However, we observed that such a procedure significantly increases the compliance costs. The lack of precise guidance at early stages of the development can lead to complications at later stages, as it becomes then harder to trace back the root causes of failures.
Moreover, developers are not necessarily equipped with the tools and methods to properly document the choices and decisions made at different stages.
Thus, the earlier the discussions between the auditor and the auditee start, the more can be taken out of the audit.
In particular, our framework delineates the procedures to be adopted gradually, enabling both an increased traceability at a later stage as well as a significant cut in the compliance cost regarding the various upcoming regulations.

Perhaps a more demanding approach would be to deeply intermingle the audit procedure with the very design and deployment the ML system.
Producers and auditors would then work hand-in-hand (or, in fact, in a slightly adversarial way, given the nature of auditing) during every phase of the development.
Such a costly procedure can obviously only be justified in high-risk applications.

\paragraph{Database of Documented AI-related Risks}

One of the impediments during ML auditing lies in the difficulty to characterise the risks incurred by the deployed system.
Thus, practices such as risk registers, widely adopted by the aviation industry, could be relevant for the ML industry.
Registering past failures and, furthermore, gathering the information necessary to understand the causes of an incident can help defining preventive measures to adopt in the future \cite{mcgregor2020preventing, shneiderman2020bridging, campbell2007evolution}. The utility of such a tool would be even greater if shared across organisations.

Useful databases such as the ``AI, Algorithmic and Automation Incident \& Controversy'' (AIAAIC) \cite{charliepow} repository exist and are useful for getting an overview of the most common and most publicised past incidents.
However, the database itself is insufficient to be used in an audit. For that, each incident would need to be documented with items such as,

\begin{itemize}
    \item failed control measures,
    \item successful control measures,
    \item the related literature on the subject,
    \item the actual impact of the incidents, including all the actors and subjects involved in the system,
    \item the estimated cost incurred by the incident,
    \item similar examples of occurrence,
    \item a detailed specification of the conditions of occurrence. Here again, one can resort to the aforementioned lifecycle model to serve as a common template when documenting these conditions.
\end{itemize}

One way to build such a database would be collaboratively. However, even single organisations would already benefit from gathering all their accumulated knowledge related to risks in a way to be re-used throughout the organisation.

\section{Conclusion and Outlook}\label{sect_conclusion}

In this paper, we have presented a procedure that takes the discussion around machine learning systems auditing towards a pragmatic direction. In particular, we have introduced a process that can be easily understood by ML practitioners, process owners, and information system auditors, hence fostering a greater adoption of auditing practices.

Our procedure mainly stands out in that it also defines a lifecycle model and a risk assessment procedure that embed ethical requirements, encouraging practices that increase the transparency and the accountability of the audited systems.
Our proposed lifecycle model explicitly emphasises the necessity to precisely specify the requirements for  the machine learning core elements, i.e., the data, the model, and the metrics, as well as the precise deployment specification (newly generated data, logs, user experience, etc.) This way, a proper quality assessment can be conducted at each iteration of the lifecycle which in turn serves the auditing downstream.

We illustrated our procedure on real-world use cases and discussed how they contributed to the definition of the audit procedure itself. Furthermore, these pilots allowed us to unveil the remaining challenges of the field.

The auditing of ML systems in the current stage heavily relies on the specifications that the audited organisation itself makes, which can be a source of blind spots. It is then imperative to define systematic ways to challenge these specifications. Future standards and regulation will likely bring a partial answer to this need. However, to account for the highly unpredictable and transformative nature of these systems, further practices are to be developed and adopted.

\begin{acks}
This work is part of the etami project (www.etami.eu).
\end{acks}

\bibliographystyle{ACM-Reference-Format}
\bibliography{bib}

\appendix

\section{Standard terminology}
\label{appendix:evaluator}
According to ISO standardization, the terms \emph{audit} and \emph{auditor} are used in the context of process assessment (ISO 17021 Conformity assessment - Requirements for bodies providing audit and certification of management systems) and the terms \emph{evaluation} and \emph{evaluator} are used in the the context of product assessment (ISO 17065 Conformity assessment - Requirements for bodies certifying products, processes and services). In our paper, even though the focus lies on product assessment, we purposely only refer to audit and auditor for the sake of simplification, clarity and conformity with the scientific literature, which is often not aware of this difference in terminology. %

\section{The different types of auditing}
\label{appendix:int_ext_audit}

Auditing procedures can be distinguished based on two dimensions. The first dimension is the relation of the auditor to the product to be audited, more specifically, to the developing team. If a member of the developing team itself is performing the audit, this is called a first-party audit (or self-assessment). If a member of a different team of the same organisation, a member of a related organisation, or a customer is performing the audit based on internal information, it is called a second-party audit. Both kinds of (internal) audits may imply a conflict of interest when it comes to making weaknesses visible. A third party (external) audit, however, avoids this problem by letting an independent auditor perform the audit\footnote{This explanation is based on ISO 17065:2012 Conformity assessment - Requirements for bodies certifying products, processes and services}.

\begin{figure}
\includegraphics[width=0.45\textwidth]{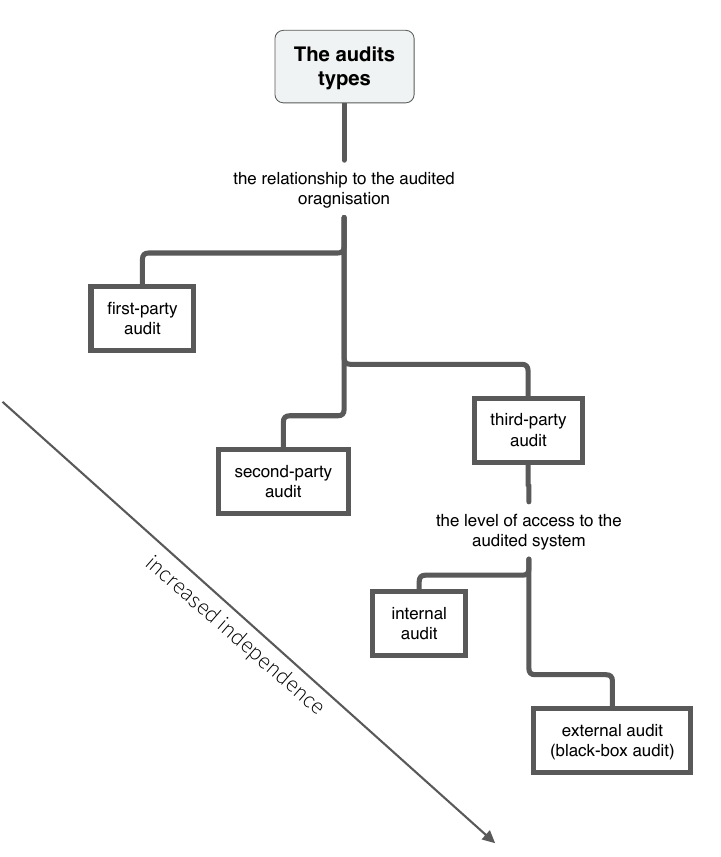}
\caption{The distinction between the different types of auditing.}
\end{figure}

The second dimension is the level of access and insight the auditor has. In case of an internal audit, the auditor can access, or at least request, any information necessary to perform a thorough audit. Obviously, every first and second party audit is also an internal audit. There are also third party audits having that kind of access, for example, when an external auditor gets the authority to access internal information during a certification program. However, if the auditor has no access to internal information, but only the same access as a regular customer or user, the product to be audited can only be examined as a black box~\cite{Diakopoulos2014}. %
As there currently is no proper terminology to differ between these two kinds of third party audits, we propose to refer to third party audits for the former and to \emph{black box audits} for the latter, as a system can also be treated as black box for internal audits.

\section{The risk assessment database}
\label{appendix:risk_assess}

In order to bootstrap the risk assessment procedure, we borrow the seven requirements of ethical and trustworthy AI as defined by the European Commission \cite{hleg2020assessment} and further relate each element of their assessment list into one or more elements of our lifecycle model and construct a database for risk assessment.

Once the relevant phases and steps are selected from the lifecycle model, one can filter easily and retain the most important questions to raise by means of a simple filter in the database.

We provide the complete list of questions at our database \footnote{\url{https://airtable.com/shrJ5z4DwmHRBoWRB/tblQuoArOc1uAiGnP}}.

\begin{figure}
\includegraphics[width=0.95\textwidth]{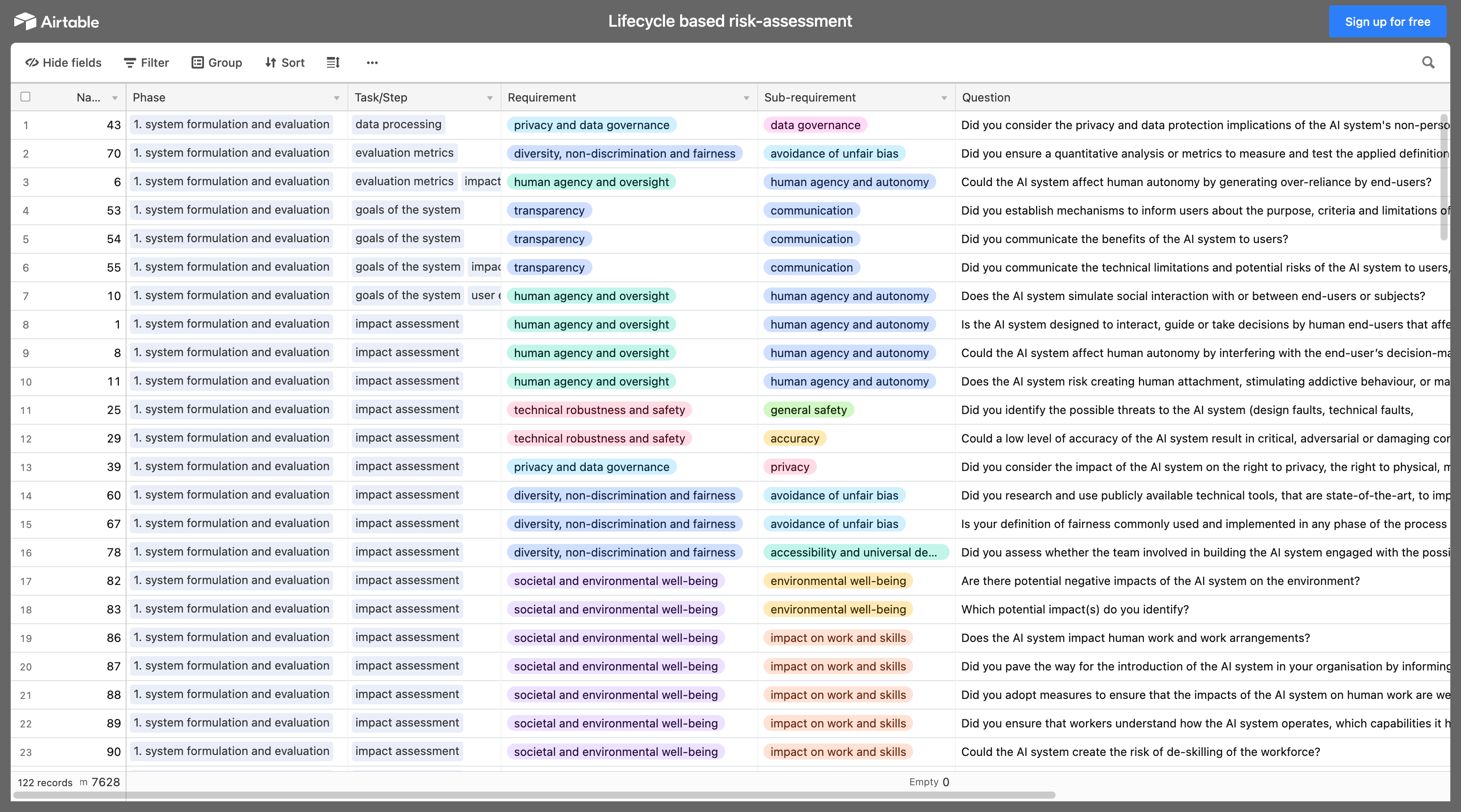}
\caption{A screenshot of the list of questions that bootstrap our proposed risk-assessment procedure.}
\end{figure}

The database we present here is mainly an example of implementation (although it can already be applied in practice). Ideally, such a database is continuously enriched as the risks of ML systems are better understood.

\section{Lifecycle mapping of Pilot 1}

\paragraph{System formulation:}

\begin{description}
    \item[Goals:] calibration of a complex safety component, which must be fully operational under a variety of environmental conditions.
    \item[Legacy systems:] not applicable for this use case
    \item[Evaluation metrics:] not known
    \item[System subjects:] safety component
    \item[System users:] domain expert
    \item[Societal context:] not relevant
    \item[User experience:] straightforward as the system is used by domain experts
    \item[Security assessment:] not relevant for this use case
    \item[Impact assessment:] non-safety critical context; privacy, transparency, explainability, and human autonomy.
\end{description}

\paragraph{Data:}

\begin{description}
	\item[Data specification:] not known
	\item[Data collection:] domain experts' historical knowledge
	\item[Data curation:] not known
	\item[Data extraction:] not known
	\item[Quality assessment:] input data are analysed by the development team and manually rejected if not complete or of low quality
\end{description}

\paragraph{Model:}

\begin{description}
	\item[Model Specification:] not known
	\item[Feature Engineering:] not known
	\item[Training/Optimisation:] not known
	\item[Validation/Interpretation:] not known
	\item[Quality Assessment:] not known
\end{description}

\paragraph{Deployment/OP:}
\begin{description}
	\item[Sandboxing:] to be conducted to test mitigation measures
	\item[Operational Logging:] missing
	\item[Continuous Testing:] conducted to asses the quality of the recommendations
	\item[Reliability Assessment:] potential over-reliance mitigation measures to be adopted
	\item[Black-box-auditing:] detection of potential ethical concern of over-reliance to be solved by conducting a randomised display of the parameters
	\item[Post-market analysis:] to be conducted once the system is launched on the market
\end{description} %

\section{Lifecycle mapping of Pilot 2: the GARMI vision module}
\label{appendix:garmi}
In this section, we detail the mapping of GARMI's vision module to the proposed lifecycle model as follows:

\paragraph{System formulation:}
\begin{description}
    \item[Goals:] recognising an elderly patient's facial expression, gestures and skeleton are crucial features for coherent and safe human robot interaction.
    \item[Legacy systems:] not applicable for this use case
    \item[Evaluation metrics:] success rates recorded during testing and validation phase.
    \item[System subjects:] elderly patient/person
    \item[System users:] health care professionals and elderly patient/person
    \item[Societal context:] elderly person's home, professional healthcare environment
    \item[User experience:] User studies are being/will be conducted to improve the system based on the stakeholders experience and feedback.
    \item[Security assessment:] In the GARMI project we aim to examine a software and hardware platform for security vulnerabilities along with project progress. This part is done with the supervision of our security partners. \cite{nside}.
    \item[Impact assessment:] input from ethics group ...
\end{description}
\paragraph{Data:}
\begin{description}
    \item[Data specification:] This is done after each stage of the learning, validation and test phase.
    \item[Data collection:] This concerns the participants' facial expressions and skeleton. This will be used for medical observations as well as engaging safety features when interacting with GARMI.
    \item[Data curation:] not yet applied
    \item[Data processing:] Data are processed offline and online depending on the application or case of use.
    \item[Data extraction:] not applicable at this
    \item[Quality assessment:] not applicable at this stage
\end{description}
\paragraph{Model:} not applicable at this stage
\paragraph{Deployment/OP:} not applicable at this stage

\end{document}